\documentclass[10pt,twocolumn,letterpaper]{article}

\usepackage{cvpr}              % To produce the CAMERA-READY version
\usepackage[dvipsnames]{xcolor}

\definecolor{cvprblue}{rgb}{0.21,0.49,0.74}
\usepackage[pagebackref,breaklinks,colorlinks,citecolor=cvprblue]{hyperref}
\usepackage{multirow}
\usepackage{times}
\usepackage{epsfig}
\usepackage{graphicx}
\usepackage{amsmath}
\usepackage{amssymb}
\usepackage{booktabs}
\usepackage{makecell}
\usepackage{multirow}
\usepackage{caption}
\usepackage{gensymb}
\usepackage[lined,boxed,commentsnumbered,ruled]{algorithm2e}
\usepackage{rotating}
\usepackage{bm}
\usepackage{colortbl}
\usepackage{color}
\usepackage{pifont}
\usepackage{makecell} 
\usepackage{wrapfig}
\usepackage{xspace}
\usepackage{hhline}

\newcommand{\yr}[1]{\textcolor{black}{#1}}

\def\paperID{1641} % *** Enter the Paper ID here
\def\confName{CVPR}
\def\confYear{2024}

\title{BA-SAM: Scalable Bias-Mode Attention Mask for Segment Anything Model}

\author{Yiran Song$^{1}$\footnotemark[1], 
Qianyu Zhou$^{1}$\thanks{\textit{The first two authors contributed equally to this work.}}, 
Xiangtai Li$^2$, 
Deng-Ping Fan$^3$,  
Xuequan Lu$^4$\footnotemark[2],
Lizhuang Ma$^1$\thanks{\textit{Corresponding author.}}
\\$^1$Shanghai Jiao Tong University; $^2$ Nanyang Technological University; \\$^3$ Nankai University; $^4$
La Trobe University \\
$^1${\tt\small \{songyiran,zhouqianyu,lzma\}@sjtu.edu.cn},   \\ $^2${\tt\small  xiangtai94@gmail.com}, 
$^3${\tt\small dengpfan@gmail.com}
$^4${\tt\small xuequan.lu@deakin.edu.au} \\
{\tt\small \textbf{Code:}\url{https://github.com/zongzi13545329/BA-SAM}}
}

\begin{document}

\twocolumn[{%
\renewcommand\twocolumn[1][]{#1}%
\maketitle
\begin{center}
    \centering
    \vspace{-13mm}
    \includegraphics[scale=0.30]{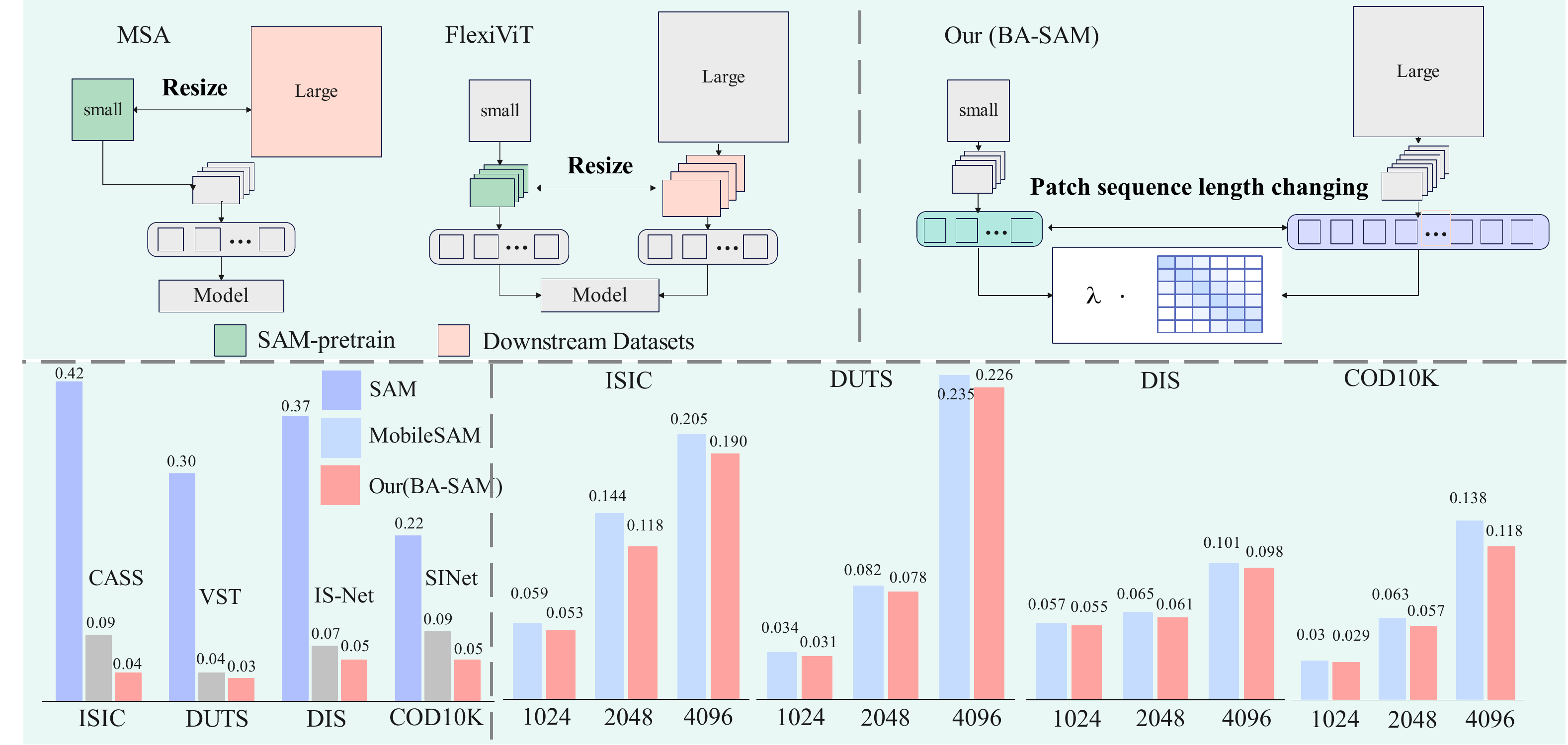}
    \vspace{-2.5mm}
    \captionof{figure}{\textbf{Top}: contrast between prior methods~\cite{zhang2023customized,beyer2023flexivit} and BA-SAM. For large-scale datasets, previous approaches often resize images or change patch sizes to handle the issue of varying resolutions. In contrast, we propose a Scalable Bias-Mode Attention Mask (BA-SAM), which enhances SAM’s adaptability to varying image resolutions while eliminating structure modifications. \textbf{Bottom (left)}: We introduce a generalized model that outperforms state-of-the-art methods across four datasets.  \textbf{Bottom (right)}: With resolution variations, prior models' performance degrades drastically. Instead, BA-SAM consistently alleviates this issue ({The evaluation metric is MAE}).
    }
    \label{teaser}
\vspace{-2mm}
\end{center}
}]

\let\thefootnote\relax\footnotetext{\textsuperscript{*}Equal contribution.}
\let\thefootnote\relax\footnotetext{\textsuperscript{\dag}Corresponding author.}

\begin{abstract}
\vspace{-4mm}
In this paper, we address the challenge of image resolution variation for the Segment Anything Model (SAM). SAM, known for its zero-shot generalizability, exhibits a performance degradation when faced with datasets with varying image sizes. Previous approaches tend to resize the image to a fixed size or adopt structure modifications, hindering the preservation of SAM's rich prior knowledge. Besides, such task-specific tuning necessitates a complete retraining of the model, which is cost-expensive and unacceptable for deployment in the downstream tasks. 
In this paper, we reformulate this issue as a length extrapolation problem, where token sequence length varies while maintaining a consistent patch size for images of different sizes. To this end, we propose Scalable Bias-Mode Attention Mask (BA-SAM) to enhance SAM's adaptability to varying image resolutions while eliminating the need for structure modifications. Firstly, we introduce a new scaling factor to ensure consistent magnitude in the attention layer's dot product values when the token sequence length changes. Secondly, we present a bias-mode attention mask that allows each token to prioritize neighboring information, mitigating the impact of untrained distant information. 
Our BA-SAM demonstrates efficacy in two scenarios: zero-shot and fine-tuning. Extensive evaluation on diverse datasets, including DIS5K, DUTS, ISIC, COD10K, and COCO, reveals its ability to significantly mitigate performance degradation in the zero-shot setting and achieve state-of-the-art performance with minimal fine-tuning. Furthermore, we propose a generalized model and benchmark, showcasing BA-SAM's generalizability across all four datasets simultaneously. 

\end{abstract}
\vspace{-6mm}    

\section{Introduction}
\label{sec:intro}
Recently, the computer vision community~\cite{20,39,he2016deep,chen2021exploring,xu2021semi,gu2021pit,zhou2022adaptive,zhou2022context,zhou2022generative,zhou2022uncertainty,zhou2022domain,zhou2023self,zhou2022transvod,he2021end,zhou2023instance,feng2022dmt,song2023rethinking,long2023rethink,long2023diverse,zhou2024est} has experienced a surge in the development of various foundation models~\cite{CLIP,align_icml,fang2023eva}. Notably, Meta has introduced SAM (Segment Anything Model)~\cite{SAM}, a \yr{prompt} model that has made a significant impact. SAM can segment any object in an image or video by incorporating a single visual prompt, such as a box or a point, without requiring additional training.
SAM is trained on an extensive SA-1B dataset~\cite{SAM}, consisting of over 11 million images and one billion masks. Its emergence has undeniably showcased robust generalization capabilities across diverse images and objects, paving the way for new possibilities and avenues in intelligent image analysis and understanding~\cite{mobile_sam,chen2023sam,zhang2023customized,ji2023segment}. Based on SAM, Some variants have been proposed, such as MobileSAM~\cite{mobile_sam} and SAM-Adapter~\cite{chen2023sam}. These efforts typically focus on improving SAM's performance on specific datasets.

During the pre-training of SAM~\cite{SAM}, the input image size is fixed at 1024. As a foundational model, SAM is expected to exhibit generalization capabilities across various downstream tasks, each associated with datasets featuring different image sizes. This is particularly crucial for high-resolution (HQ) datasets characterized by larger dimensions and more details.
SAM performs well when the resolutions align with its training resolution of 1024. However, significant performance degradation is observed when inferring with resolutions larger than 1024. Hence, we aim to study a practical and realistic problem to enhance  SAM's adaptability to varying image resolutions of different datasets.

Since SAM adopts the standard Vision Transformer~\cite{dosovitskiy2020image} architecture, there are two common approaches to address the inconsistency between training and inference sizes for the ViT architecture. As depicted in Fig.~\ref{teaser}, the first approach, \emph{e.g.,} MSA~\cite{zhang2023customized} and SAM-Adapter~\cite{chen2023sam}, involves directly resizing all datasets to match a predefined size.
Conversely, the second approach, exemplified by FlexiViT~\cite{beyer2023flexivit}, entails adjusting the patch size to accommodate larger image resolutions. Nevertheless, tuning the image or patch size necessitates a complete retraining of the model, which is cost-expensive and unacceptable for deployment in the downstream tasks. Besides, it
prevents leveraging the rich prior knowledge reserved in the pre-trained model of  SAM. 
As such, our objective is to explore a solution that enhances SAM's adaptability to datasets of varying resolutions while avoiding structural modifications to SAM.

In this paper, we introduce a novel perspective that reframes the challenge of image resolution variation as a length extrapolation problem. Specifically, as depicted in Fig.~\ref{teaser}, for images of varying sizes, we employ different token sequence lengths while keeping a consistent patch size.
It has been observed that the inconsistency in token length between training and prediction is a key factor in performance degradation. This inconsistency manifests in two aspects: 
Firstly, changes in token length lead to variations in the magnitude of attention module values. When the dot product result becomes significantly large in magnitude, it can drive the subsequent Softmax layer into regions with minimal gradients. Consequently, the attention distribution after Softmax becomes highly concentrated, giving rise to the issue of vanishing gradients.
Secondly, longer predictions rely on untrained information, such as additional position encodings. The introduction of untrained parameters brings a considerable amount of noise to the model, which, in turn, affects its performance.

To address these issues, we propose a Scalable Bias-Mode Attention Mask (BA-SAM) to enhance the length extrapolation capability of SAM. Our approach introduces two novel designs.
Firstly, to ensure consistency in the attention layer's dot product value, we present an improved scaling factor. This factor effectively regulates the magnitude of values within the attention layer, mitigating disruptive effects resulting from substantial changes in dot product operations and context length.
Secondly, with a focus on maintaining consistency in attention focus areas, we introduce a novel bias-mode attention mask. This attention mask penalizes attention scores between distant query-key pairs, with the penalty increasing as the distance between the key and query grows. Consequently, when the context length varies, the influence of untrained distant information on each token diminishes. We achieve this mask by adding a bias after the query-key dot product, and this design is highly lightweight and could be seamlessly integrated into SAM-based models with minimal computational overhead.

Our approach demonstrates efficacy in two scenarios: zero-shot and fine-tuning. Extensive evaluations on datasets from five diverse tasks are conducted, including DIS5K~\cite{DIS}, DUTS~\cite{DUTS}, ISIC~\cite{codella2018skin}, COD10K~\cite{COD}, and COCO~\cite{lin2014microsoft}. These datasets vary in resolutions, mostly exceeding SAM's default resolution of 1024.
In the zero-shot setting, our BA-SAM alleviates the model's performance degradation caused by expanding the inference resolution without requiring any additional training. With a few fine-tuning epochs on downstream tasks, our BA-SAM consistently achieves state-of-the-art accuracy across all datasets.
Additionally, to further demonstrate BA-SAM's generalizability, we propose a generalized model and a new benchmark, which utilize one model to attain state-of-the-art performance across all four datasets simultaneously.

%-------------------------------------------------------------------------

\section{Related Work}
\label{sec:related_work}

\noindent
\textbf{Visual Foundation Models.} Models that are trained on broad and can be adapted to numerous downstream tasks are called ``Foundation Models''~\cite{bommasani2021opportunities,wang2023large,liang2022foundations,li2023transformer,wu2023open,yuan2024ovsam,zhou2023edgesam,li2024omg}. These models, Vision-Language Models (VLM) (CLIP~\cite{CLIP} and DALL-E~\cite{ramesh2022hierarchical}) combine computer vision and natural language processing to understand and generate descriptions or analyze visual content using textual and visual information. Masked Image Modeling~\cite{xie2022simmim,liu2022swin} (MIM) masks parts of an image during the training to encourage a model to learn contextual information and complete missing regions. SAM~\cite{SAM} is a model designed for segmenting objects or areas in images, offering precise segmentation capabilities.
% % In the experimental section, we utilize a variant of SAM called mobileSAM ~\cite{mobile_sam} as the baseline method. Nevertheless, our design optimizes the Transformer itself, making it applicable to all Transformer-based Vision foundation models. 
% % Qianyu
We use a variant of SAM called MobileSAM~\cite{mobile_sam} as the baseline method. 
% Since our design optimizes the Transformer itself, it can be generalized to all Transformer-based models for scale variant problems.

% it applies to all Transformer-based vision foundation models. 

\begin{figure*}[t!]
\centering
\includegraphics[width=\textwidth]{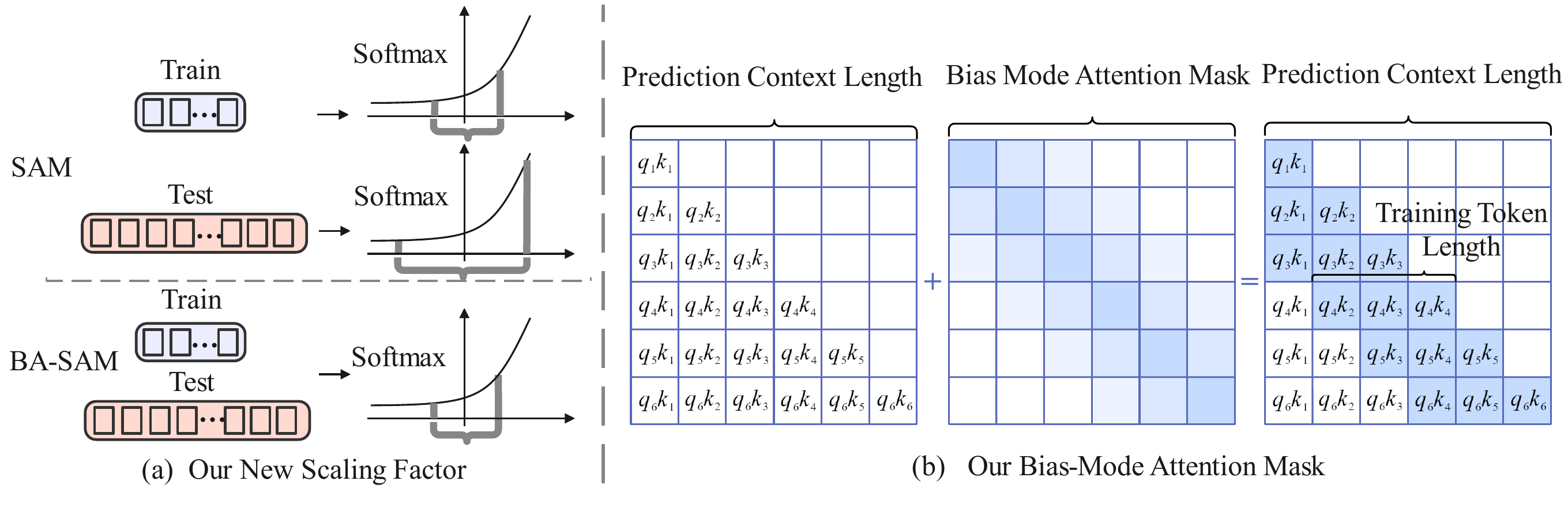}
% \vspace{-7mm}
\caption{ 
Illustration of the proposed BA-SAM method. (a) In the original SAM, when input token sequences length varies during testing, the magnitude of the Softmax outputs changes drastically. We propose a new scaling factor to address this issue. (b) We introduce a bias-mode attention mask, which produces increasing penalties on attention scores as the distance between the query and key grows. 
}
\label{fig:sum}
% \vspace{-3mm}
\end{figure*} 

\noindent
\textbf{Resolution Variation Processing.}
To enable models to be more adaptable to variations in resolutions, traditional methods to deal with VIT have relied on adjustments to positional embeddings~\cite{lee2023pix2struct} and patch sizes~\cite{beyer2023flexivit,8,18,63,34_flex,20,31_felex}. Patch n' Pack~\cite{dehghani2023patch} employed sequence packing during the training to handle inputs with arbitrary resolutions and aspect ratios. All of them necessitate training from scratch, incurring substantial computational and time costs. In contrast to previous approaches, we extend the concept of \textit{length extrapolation} from NLP into the context of addressing scale variations in CV. Length extrapolation refers to a model's ability to generalize well to longer inputs than those it was trained on. In NLP, it has been successfully used, such as in ALIBI~\cite{press2021train} and KERPLE~\cite{chi2022kerple}, to enable models to adapt to longer sequences without significant performance degradation. Our approach seamlessly extends to two scenarios: zero-shot and finetuning, allowing us to leverage prior knowledge embedded in the SAM and significantly reduce training efforts.

\noindent
\textbf{Parameter Efficient Tuning.}
There have been some pioneering works for the Parameter Efficient Tuning (PEFT) of visual models, such as AdaptFormer~\cite{chen2022adaptformer} and visual prompt tuning (VPT)~\cite{jia2022visual}. He et al.~\cite{He_Zhou_Ma_Berg-Kirkpatrick_Neubig_2021} analyzed the unified view among PETL techniques such as prefixtuning~\cite{Li_Liang_2021}, Prompt-tuning~\cite{jia2022visual}, and adapter~\cite{chen2022adaptformer}. Our method belongs to the category of Parameter Efficient Tuning.

\noindent
\textbf{Visual Attention Modeling.}
Various studies have incorporated attention mechanisms into neural network architectures designed for visual tasks~\cite{35,36,37,38,ji2023sam,9789317}. These mechanisms are employed in a channel-wise manner to capture cross-feature information~\cite{39,40,41}. They are also used for selecting paths in different branches of a network~\cite{42}, or a combination of both strategies~\cite{43}. 
% For instance, squeeze-and-excite network~\cite{44} features an attention-like module for modeling channel-wise relationships within layer features. Li~\emph{et al.}~\cite{37} utilize the attention mechanism to adapt the receptive field of neurons between network branches. 
The advent of transformers has led to hybrid architectures that integrate other modules. Bello's work~\cite{45} introduces approximate content attention with a positional attention component. Child \emph{et al.}~\cite{23} observe that many early layers in the network learn locally connected patterns akin to convolutions, indicating that hybrid architectures inspired by both transformers and convolutional networks are a compelling design choice. Several recent studies explore this approach for various tasks~\cite{46,47,48,graham2021levit}. 
In contrast to prior work, we do not introduce a new attention structure. Instead, we offer theoretical proof for optimizing existing attention mechanisms. This resulting optimization approach is applicable across various attention designs and demonstrates strong performance across multiple datasets.

\section{Preliminaries}

\noindent
\textbf{SAM.}
Segment Anything Model (SAM)~\cite{SAM} consists of three core modules: image encoder, prompt encoder, and mask decoder. 
It has been trained on SA-1B dataset~\cite{SAM}, which comprises more than 1 billion automatically generated masks. 
Consequently, SAM exhibits valuable and robust zero-shot generalization to new data without necessitating further training. further details can refer to~\cite{SAM}. Our Scalable Bias-Mode Attention Mask (BA-SAM) focuses its optimization on the image encoder while keeping the structures of the mask decoder and prompt encoder unchanged.

\noindent
\yr{\textbf{Attention in Transformer.}
In this work, we define the input sequence of image patches, $\mathbf{x} = (\mathbf{x}_1, \dots, \mathbf{x}_n)$ with length $N$, where $\mathbf{x}_i \in \mathbb{R}^{d_x } $. $q_i$, $k_j$,$v_j$ are calculated by $\mathbf{x}_i \mathbf{W}^Q$ , $\mathbf{x}_j \mathbf{W}^K$, $\mathbf{x}_j \mathbf{W}^V$. Here, the projections $\mathbf{W}^Q, \mathbf{W}^K, \mathbf{W}^V \in \mathbb{R}^{d_x \times d_k}$ are parameter matrices.}

\noindent \textbf{(i) Scaling Factor.} The two most commonly used attention functions are additive attention~\cite{bahdanau2014neural} and dot-product attention~\cite{Vaswani2017Attention}.
% \lxt{Please define x, q, k before using them.} 
The vanilla Transformer chooses dot-product attention for its space efficiency in practice. However, for larger values of $d_k$, the dot products grow large in magnitude, pushing the Softmax function into regions with minimal gradients. They use \emph{scaling factor} $\lambda_d=\frac{1}{\sqrt{d_k} }$ to scale the dot products, where $d_k$ denotes the dimension. To better analyze the role of scaling factor, we express the output element $\mathbf{O}_{i}$ and the weight coefficient $a_{i,j}$ as follows: 
    \begin{align}
    \mathbf{O}_{i} &= \sum_{j=1}^{N}a_{i,j}v_{j},  \quad
    a_{i,j} = \frac{e^{\lambda_d q_i \cdot k_j}}{\sum_{j=1}^{N}e^{\lambda_d q_i \cdot k_j}}, \label{attention}
\end{align}
% \end{equation}
where $\lambda_d$ represents the scaling factor. %independent of $Q$ and $K$.

\noindent \textbf{(ii) Absolute \& Relative Position Encoding.}
\label{position}
The original Transformer~\cite{Vaswani2017Attention} incorporates absolute non-parametric positional encoding $p=(p_1,\dots,p_n )$ with $x$ as $x_i=x_i + p_i$. Other works replace them with parametric encoding~\cite{gehring2017convolutional} or adopted Fourier-based kernelized versions~\cite{parmar2018image}. Absolute position encoding enforces a fixed size for inputs. Recent work ~\cite{shaw2018self} considers the pairwise relationships between elements, which encodes the relative position between input $x_i$ and $x_j$ into vectors $p_{i,j}^v,p_{i,j}^q,p_{i,j}^k \in \mathrm{R}^{d_k} $. Then, we reformulate Eq. \eqref{attention} as follows: 
\begin{equation}
    \mathbf{O}_{i} = \sum_{j= 1}^{N}a_{i,j}\left(v_{j}+ p_{i,j}^v \right), 
\end{equation}
\begin{equation}
    a_{i,j}=\frac{e^{\lambda \left(q_i+ p_{i,j}^q \right)  \cdot \left (k_j+ p_{i,j}^k \right)  }}{\sum_{j=1}^{N}e^{\lambda \left(q_i+ p_{i,j}^q \right)  \cdot \left (k_j+ p_{i,j}^k \right)  }},
\end{equation}
where $p_{i,j}^v,p_{i,j}^q,p_{i,j}^k $ is learned during training.

\section{Methodology}
\label{method}
\yr{Based on the preliminaries, we further analyze the characteristics of SAM: the original SAM sets the input to a fixed resolution of 1024, where it uses absolute position coding and the dot product. Thus, there are significant limitations in the processing of length extrapolation problems.} To address this, \yr{as shown in Fig.~\ref{fig:sum},} we present a Scalable Bias-mode Attention Mask (BA-SAM). In Sec.~\ref{sec:4.1}, we provide a theoretical explanation for the scaling factor used in the original Transformer and introduce a new scaling factor to regulate the magnitude inconsistency caused by length extrapolation. In Sec.~\ref{sec:4.2} we design a bias-mode attention mask to place more focus on neighboring tokens, mitigating the impact of untrained distant information.
Finally, we explain how we will embed our BA-SAM into the SAM-based structure in Sec.~\ref{sec:4.3}.

\subsection{New Scaling Factor}
\label{sec:4.1}

We observe that in the original attention module of SAM~\cite{Vaswani2017Attention} when the dot product becomes significantly large in magnitude, it can drive the Softmax layer into regions with minimal gradients. This is because the attention distribution after Softmax becomes highly concentrated, giving rise to the issue of vanishing gradients.
Upon closer examination of Eq. \eqref{attention}, it is obvious that the computation of the $q \cdot k$ %$QK$ 
the term is intrinsically tied to both the token sequences length $N$ and the dimension $d_k$. 
When the token sequence length $N$ and the dimension $d_k$ significantly increase, the overall efficacy of the attention is affected, thus leading to a noticeable performance degradation.

To address this issue, we attempt to design a new scaling factor that allows the model to cope with variations in $N$ and $d_k$.  When $N$ or $d_k$ grows significantly, we expect to regulate the magnitude of the values within the attention layer, maintaining a similar magnitude. \cite{Vaswani2017Attention} introduced a scaling factor $\lambda=\frac{1}{\sqrt{d_k}}$ to counteract the effect of the large growth in magnitude due to the dot products. Below we will provide a theoretical derivation of this scaling factor, and then elaborate on our proposed new scaling factor. 

\noindent \textbf{The dimension $d_k$.} \yr{Following work \cite{Vaswani2017Attention}, we assume that the components of $q$ and $k$ are independent random variables with mean 0 and variance 1.} 
The mean of $q\cdot k$ is:
\begin{equation} \label{4}
\begin{aligned}
\mathrm{E}[q \cdot k] & =\mathrm{E}\left[\sum_{i=1}^{d_k} q_i k_i\right] =\sum_{i=1}^{d_k} \mathrm{E}\left[q_i\right] \mathrm{E}\left[k_i\right] =0
\end{aligned}
\end{equation}
Similarly, we formulate the variance of $q\cdot k$ as follows:
\begin{equation}\label{5}
\begin{aligned}
\operatorname{var}[q \cdot k] & =\operatorname{var}\left[\sum_{i=1}^{d_k} q_i k_i\right] =\sum_{i=1}^{d_k} \operatorname{var}\left[q_i\right] \operatorname{var}\left[k_i\right] =d_k
\end{aligned}
\end{equation}

Given this, we can approximately consider the $q\cdot k$ values to be within the range of $-3\sqrt{d_k} $ to $3\sqrt{d_k}$, according to properties of Gaussian distribution. For larger models, $d_k$ is generally a larger positive value, resulting in a significant increase in the magnitude of numerical values of $q\cdot k$, compared to the additive attention option, which has the range of $\left[-3, 3\right]$. Consequently, %the attention distribution closely resembles a one-hot distribution. 
the attention distribution after Softmax becomes highly concentrated. 
This leads to severe gradient vanishing, which hampers the effectiveness of the training and can lead to less desired performance. 
As the $q\cdot k$ values lie in the range of $[-3\sqrt{d_k}, 3\sqrt{d_k}]$, the scaling factor can be simply defined as $\lambda_d = \frac{1}{\sqrt{d_k}}$, in order to maintain a similar magnitude.

\noindent \textbf{Our new scaling factor.} 
We have provided the interpretation of how the original scaling factor was designed. Now, we explain the design of our new scaling factor.

According to Eq. \ref{4} and Eq. \ref{5}, the scale of $q\cdot k$ has been consistent with the additive attention by $\lambda_d$, which can be seen as $a_{i,j}$ is independent of $d_k$. 
We simplify $\lambda_d  q_i\cdot k_j$ into $x_{i,j}$ and further discuss the effect of length $N$ on $a_{i,j}$.

In Eq.\ref{attention} , $a_{i,j}$ can be seen as the conditional distribution with $i$ as the condition and $j$ as the random variable. Inspired by \cite{kexuefm}, we introduce information entropy to constrain $a_{i,j}$. Specifically, entropy is a measure of uncertainty, and we expect the uncertainty of $a_{i,j}$ to be insensitive to the length $N$ (\emph{i.e.,} the value of each $a_{i,j}$ will change when the token increases, but the entropy value of the overall $a_{i,j}$ can remain relatively stable).
The entropy of $a_{i,j}$ is :$\mathcal{H}_i=-\sum_{j=1}^N a_{i, j} \log a_{i, j}$  
and we substitute Eq.~\eqref{attention}:

\begin{equation} \label{entropy}
\mathcal{H}_{i} = \log \sum_{j=1}^{N} e^{\lambda_n x_{i,j} }-\frac{\sum_{j=1}^{N} e^{\lambda_n x_{i,j} }\left(\lambda_n x_{i,j} \right)}{\sum_{j=1}^{N} e^{\lambda_n x_{i,j} }}
\end{equation} 
Then, we substitute the approximate estimates into Eq. (\ref{entropy}):
\begin{equation}
\begin{aligned}
& \sum_{j=1}^N e^{x_{i,j}}=N \times \frac{1}{N} \sum_{j=1}^N e^{x_{i,j}} \approx N \mathbb{E}_j\left[e^{x_{i,j}}\right] \\
& \mathbb{E}_j\left[e^{x_{i,j}}\left(x_{i,j}\right)\right] \approx 0, \mathbb{E}_j\left[e^{x_{i,j}}\right] = O(1) 
\end{aligned}
\end{equation}
% and $ \mathbb{E}_j\left[e^{x_{i,j}}\left(x_{i,j}\right)\right] \approx 0$,$\mathbb{E}_j\left[e^{x_{i,j}}\right] = O(1) $ . 
We wish to use $\lambda_n$ to offset the effect of $N$ on $\mathcal{H}_i$. Then, we have the following result: 

\begin{equation} \label{lamda}
\mathcal{H}_i \approx \log N- k \lambda_n = 0 \Rightarrow \lambda_n=\frac{\log N}{k}
\end{equation}
where $k$ is a parameter value. We denote the token sequence length during the training as $N_{train}$ and the token sequence length during the testing as $N_{test}$, where $N_{test} >> N_{train}$. When $N =N_{train} $, $\lambda_n = 1 $ (consistent with the training length). As such, $k=\log N_{train} $ and finally we have $ \lambda_n = log_{N_{train}}N_{test}$.
Considering both $\lambda_d$ and $\lambda_n$, we can ultimately derive our new scaling factor as: 
\begin{equation}
    \lambda = \lambda_d  \lambda_n = \frac{log_{N_{train}}N_{test}}{\sqrt{d_k}}
    \label{lambda}
\end{equation}
Our new scaling factor in Eq. \eqref{lambda} ensures attention computation remains consistent, regardless of variations in $d_k$ and $N$. It will enhance the extrapolative capacity of the model.

\subsection{Bias-Mode Attention Mask}
\label{sec:4.2}

Another challenge is that changes in token sequence length will lead to variations in positional encoding. It is important to ensure the insensitivity of the model when such positional encoding variations occur during the testing.  

One possible way is absolute encoding without trainable parameters, such as Sinusoidal~\cite{Vaswani2017Attention}. It requires the position encoding to have strong local-to-global inference capabilities. 
Nevertheless, this assumes that the given function has high-order smoothness (higher-order derivatives exist and are bounded). Commonly-used positional encodings are often combined with trigonometric functions
These methods fail to satisfy the requirement of bounded high-order derivatives, making it less accurate to estimate the extrapolated results. Another potential approach is using local attention \cite{liu2021swin}, which constrains the model's field of view and remains insensitive to variations in token sequence length. However, local attention is typically implemented using a local window, necessitating modifications to the SAM structure, which requires re-training from scratch and is unacceptable for deployment 
in the downstream tasks. 
% \lxt{Please merge two paragraphs into one}

To this end, we propose enabling the attention layer to focus more on the current token's neighboring tokens.
As such, even with an increase in the length of a token sequence, each token is scarcely affected by the untrained tokens from distant positions. 
In particular, we design a simple yet effective bias-mode mask. We introduce a bias after the query-key dot product. 

\begin{figure}[t!]
\centering
\includegraphics[width=0.45\textwidth]{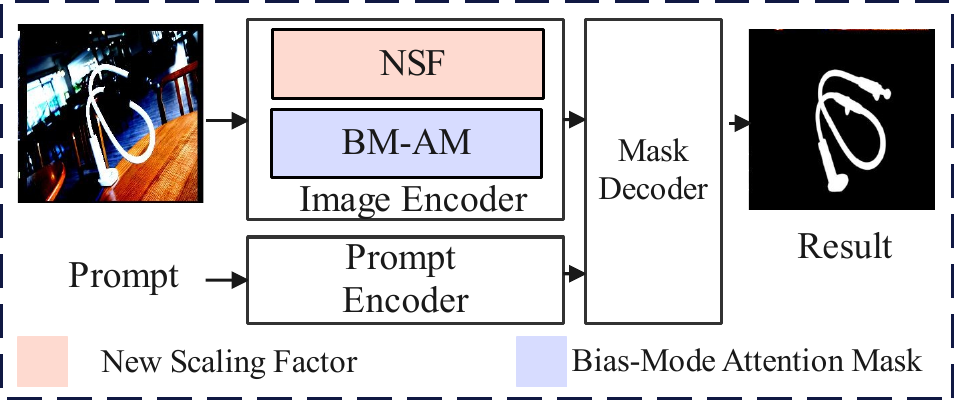}
% \vspace{-4mm}
\caption{ 
Embedding of our BA-SAM into a SAM backbone. NSF indicates our new scaling factor, and BM-AM denotes our designed bias-mode attention mask.  
}
% \vspace{-3mm}
\label{pipeline}
\end{figure}

As shown in Fig.~\ref{pipeline}, this mask exhibits a bias specified on the distance between the query-key pairs (i.e., $q\cdot k$). We expect that this proposed mask imposes penalties on attention scores between distant query-key pairs, and the penalty increases as the distance between a key $q$ and a query $k$ grows. To this end, we simply define the bias as the form of $b_{i,j} = \beta|i-j|$. 
\begin{equation}
        a_{i,j}=\frac{e^{\lambda (q_i \cdot k_j+b_{i,j}) }}{\sum_{j=1}^{N}e^{\lambda (q_i \cdot k_j+b_{i,j}) }},
\label{mask}
\end{equation}
where $\beta$ is a head-specific slope. 

We further discuss the setting of $\beta$ based on different cases. When conducting zero-shot generalization without fine-tuning, we set $\beta$ to a static, non-learned fixed value. 
The experimental section will discuss the specific value setting (Sec.~\ref{sec:exp}). When fine-tuning is required, we make $\beta$ trainable. 
Since our Bias-Mode Attention Mask is lightweight relative to the model structure, it incurs negligible training cost overhead.

\begin{figure*}[ht!]
\centering
\includegraphics[width=\textwidth]{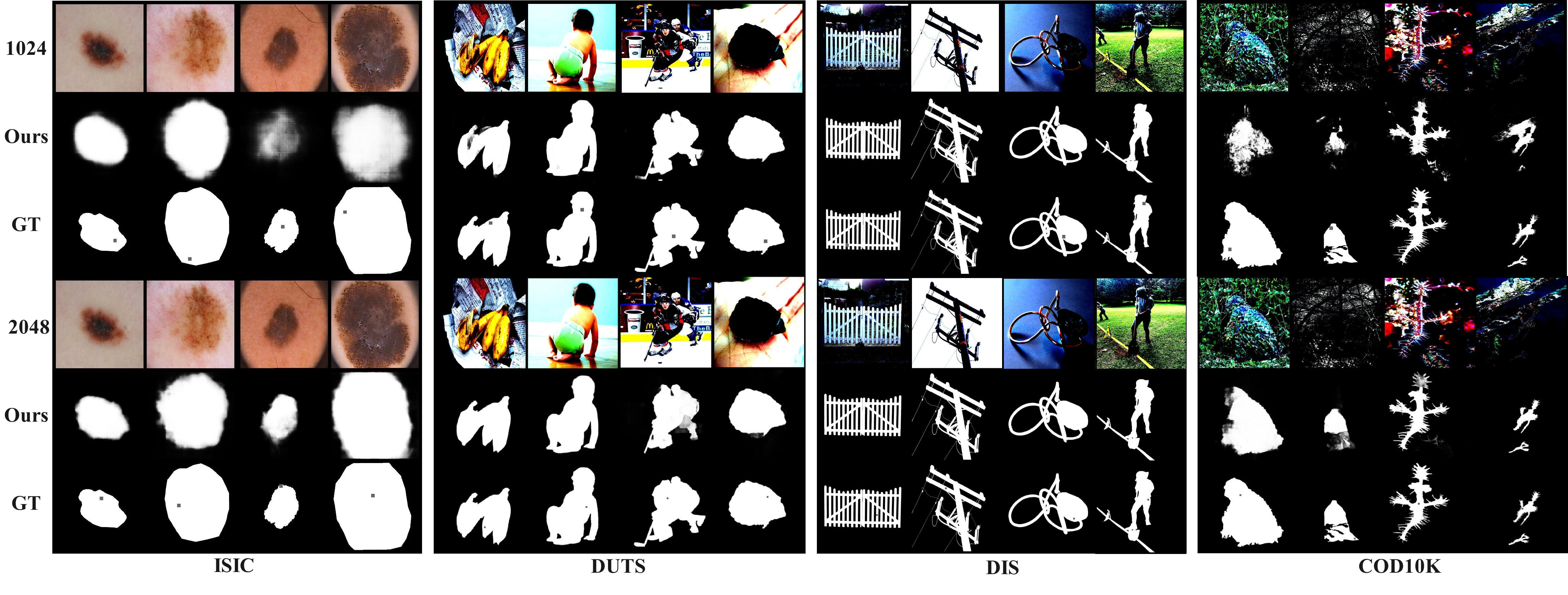}
% \vspace{-8mm}
\caption{ 
Visualization results of our BA-SAM on four object segmentation tasks, \emph{i.e.,} 
skin lesion segmentation, salient object segmentation, complex object segmentation,  camouflaged object detection,  which corresponds to four datasets:
ISIC~\cite{codella2018skin}, DUTS~\cite{DUTS}, DIS-TE4~\cite{DIS}, and COD10K~\cite{COD}. Our BA-SAM can handle the issue of varying image resolutions and segments accurately in different tasks. 
 }
\label{fig:visulazition_main}
% \vspace{-1mm}
\end{figure*}

\subsection{BA-SAM Model}
\label{sec:4.3}
As shown in Fig.~\ref{pipeline}, our BA-SAM is simple to implement, %independent of model structures, 
and can be seamlessly integrated into SAM~\cite{SAM} and its variants. Specifically, our design involves a new scaling factor (NSF) for the attention layer and a bias-mode attention mask (BM-AM). Our method does not involve any alterations to the model structure and is suited to both fine-tuning and non-fine-tuning cases. As for the fine-tuning case, it introduces negligible computational overhead as BM-AM incurs a very small amount of computation.  
\section{Experiments}
\label{sec:exp}
\begin{table*}[ht]
\centering
\resizebox{0.90\textwidth}{!}{
\begin{tabular}{r | c  c|  c c  c c  c c  c c }
\toprule[0.2em]
Method  & \begin{tabular}{c} Train\\Size\end{tabular} & \begin{tabular}{c} Test\\Size\end{tabular}  & ISIC~\cite{codella2018skin} & $\Delta$\textit{diff} & DUTS~\cite{DUTS} & $\Delta$\textit{diff} & DIS-TE4~\cite{DIS} & $\Delta$\textit{diff} & COD10K~\cite{COD} & $\Delta$\textit{diff}   
 \\
\toprule[0.2em]
 \multicolumn{11}{c}{Without fine-tuning}\\
 \toprule[0.2em]
\multirow{2}{*}{SAM~\cite{SAM}} & - & 1024 & 0.421 & -   &     0.298 &  -  & 0.362   &  -   &  0.217 & - \\
                     &      -              & 2048 & 0.601 &  18.0\%  &   0.360  &   6.2\% &  0.411  &   \underline{4.9\%}  & 0.391  & 17.4\% \\
\hline
% \!\cellcolor{gray!9.0}
\multirow{2}{*}{Ours (w~\cite{SAM})}    & - & 1024 & \textbf{0.417} &  -  &  \textbf{0.294}   & -&   \textbf{0.356} & -    &  \textbf{0.208}  & -  \\
                           &         -      & 2048 & 0.589 & 17.2\%  &  0.348   &\underline{5.4\%} &  0.406  &  5.0\%   & 0.387  & 17.9\%  \\
\hline
\multirow{3}{*}{MobileSAM~\cite{mobile_sam}} & - & 1024 & 0.463 & -  &  0.502   & - &  0.544  &  -   & 0.465  & -  \\
                           &     -               & 2048 & 0.641 &   17.8\% &  0.437   & 6.5\%  & 0.427   &  11.7\%   & 0.346  &  \underline{11.9\%}  \\
                           &                            & 4096 &  0.693 &  23.0\%  &  0.328   &  17.4\% & 0.355   &   18.9\%  &  0.300 &16.5\% \\
\hline
\multirow{3}{*}{Ours (w~\cite{mobile_sam})}       & - & 1024 & 0.452 &  -  &  0.486   & -&   0.515 & -    &  0.440  & - \\
                           &    -                        & 2048 & 0.611 &  \underline{15.9\%}  &  0.413   &7.3\% &  0.406  &  10.9\%   & 0.321  & \underline{11.9\%} \\
                           &                            & 4096 & 0.657  &  20.5\%  & 0.283
                           & 20.3\% & 0.361   &  15.4\%   &  0.246 & 19.4\% \\
\toprule[0.2em]
\multicolumn{11}{c}{With fine-tuning}\\
\toprule[0.2em]
\multirow{5}{*}{MobileSAM~\cite{mobile_sam}} 
                           & \multirow{3}{*}{1024}  & 1024 & 0.059 & -   &  0.034   & -  & 0.057   & -    & 0.030  &  - \\
                           &                            & 2048 & 0.144 &  8.5 \%  &   0.082  & 4.8\% &  0.065  & 0.8\%    & 0.063  & 3.3\%\\
                           &                            & 4096 & 0.205 &  14.6 \%  &   0.235  & 20.1\%  &  0.101  &   4.4\%  &   0.138 & 10.8\% \\
                           & \multirow{2}{*}{2048} & 2048 & 0.083 &  -  &  0.045   &  -  &  0.056  &  -   & 0.036  & -  \\
                           &                            & 4096 & 0.227 &  14.4\%  &  0.091  & 4.6\% &  0.066  &   1.0\%  &  0.059 & 2.3\%\\
\hline
\multirow{5}{*}{Ours (w ~\cite{mobile_sam})}     & \multirow{3}{*}{1024} & 1024 & \textbf{0.053} &  -  &  \textbf{0.031}   & - &  0.055  &  -   &  \textbf{0.029} & - \\
                           &                            & 2048 & 0.118 & \underline{6.5 \%}   &   0.078  & \underline{4.4\%}  &  0.061  &   \underline{0.6\%}  & 0.057  & 2.5\% \\
                           &                            & 4096 & 0.190 &  13.7\%  &  0.226   &  19.2\% & 0.098   &  4.3\%   &   0.118 & 8.6\%\\
                           & \multirow{2}{*}{2048} & 2048 & 0.080 & -   &   0.043  & - & \textbf{0.053}   &  -   & 0.033  &  - \\
                           &                            & 4096 & 0.214 &  13.4\%  &0.088   & \underline{4.4\%} & 0.061  & 0.8\%    &  0.056 & \underline{2.3\%} \\
\bottomrule[0.2em]
\end{tabular}
}
%}
% \vspace{-3mm}
\caption{ Performance comparisons in varying image resolutions. We employed the widely-used  MAE (Mean Absolute Error) score. Lower MAE scores indicate better model performance. $\Delta$\textit{diff} denotes the performance degradation due to resolution changes. Compared to the SAM~\cite{SAM} and MobileSAM~\cite{mobile_sam} baselines, our proposed BA-SAM achieves smaller degradation when encountering token sequence length changes. The best MAE performance is highlighted in bold, and the smallest performance degradation is underlined. %a tilde on each resolution of each dataset. 
}
\label{Methods}
% \vspace{-3mm}
\end{table*}

% \yr{We first describe the datasets and implementation details in Sec.~\ref{dataset}. Then, we compare BA-SAM's performance with state-of-the-art methods across multiple tasks in Sec.~\ref{results}. Finally, extensive ablation studies and analysis are conducted in Sec.~\ref{sec:ablation}. to demonstrate the effectiveness.}

\subsection{Datasets and Implementations} \label{dataset}
\noindent \textbf {Datasets.} For a comprehensive evaluation of BA-SAM, we conduct extensive experiments on a wide range of segmentation tasks, \emph{i.e.,} salient object segmentation~\cite{SIP,fan2018salient,fan2022salient}, complex object segmentation~\cite{fan2022rething}, skin lesion segmentation~\cite{9789317}, camouflaged object detection~\cite{fan2021concealed}, which correspond to four datasets: DUTS~\cite{DUTS}, DIS-TE4~\cite{DIS},  ISIC~\cite{codella2018skin} and COD10K~\cite{COD}. Besides, we verify it on the challenging COCO~\cite{lin2014microsoft} instance segmentation benchmark. \textit{More details are referred to in the supplementary material.}

\noindent \textbf{Implementation Details.}
\noindent
In zero-shot settings that do not require fine-tuning, we use the original SAM~\cite{SAM} backbone. For fine-tuning scenarios, we employ MobileSAM~\cite{mobile_sam} as the backbone. MobileSAM \cite{mobile_sam} is a SAM variant with a structure similar to ViT-Tiny~\cite{wu2022tinyvit}, and further details can be found in~\cite{mobile_sam}. MobileSAM~\cite{mobile_sam} uses a ViT-H-based SAM as a teacher network for distillation, ultimately achieving competitive accuracy compared to the original SAM but with significantly fewer parameters. 
For various object segmentation tasks, a random point is extracted from the ground truth as the prompt input during the fine-tuning phase. 
For instance segmentation, we use the ViT-B~\cite{dosovitskiy2020image} backbone and the state-of-the-art detector Deformable-DETR ~\cite{zhu2020deformable} trained on the COCO~\cite{lin2014microsoft} dataset with Swin-L~\cite{liu2021swin} backbone as box prompt generator.
\textit{More details are provided in the supplementary material.} 
% The code will be released at this \href{https://github.com/zongzi13545329/BA-SAM.git}{link}.

\noindent \textbf{Evaluation metrics.} In the experiments, we use the widely used Mean Absolute Error (MAE) and Average Precision (AP) for evaluation. A lower MAE score and a higher AP score indicate better model performance.

\subsection{Results} \label{results}
\noindent \textbf{Results of Various Object Segmentation Tasks:}
Tab.~\ref{Methods} demonstrates the effectiveness of our approach across four diverse segmentation datasets.  $\Delta$\textit{diff} denotes the value of the performance degradation due to resolution changes during the inference. The upper and lower parts of the table indicate the results without and with fine-tuning. The best MAE performance is highlighted in bold, and the smallest degradation is underlined. %a tilde on each resolution of each dataset. 
We have three observations: Firstly, our proposed BA-SAM consistently outperforms both SAM~\cite{SAM} and MobileSAM~\cite{mobile_sam} baselines on all four datasets. This is mainly because these baselines do not consider the issue of varying image resolutions. In contrast, our presented scaling factor and bias-mode attention mask explicitly handle this issue and further alleviate the performance degradation. Secondly, when testing on higher resolutions than the training size, SAM~\cite{SAM} and MobileSAM~\cite{mobile_sam} baselines show less desirable results than the original image size. In contrast, our BA-SAM incurs significantly less performance drop in different datasets.
Thirdly, during the experiment, we observe negligible computational overhead, whether fine-tuning is applied or not, which supports the claim in the method section. See Sec.~\ref{efficiency} for details.

\noindent \textbf{Results of Instance Segmentation:}
In Tab.~\ref{coco}, we evaluate the performance of our method on the COCO~\cite{lin2014microsoft} instance segmentation benchmark. For a fair comparison, all experiments are conducted in a zero-shot manner, with the same initialization parameters for the comparative methods and without the use of any additional training data. Our BA-SAM consistently outperforms SAM~\cite{SAM} and MobileSAM~\cite{mobile_sam} baselines, demonstrating better zero-shot generalization capability on instance segmentation.

\noindent \textbf{Comparisons with State-of-the-Art Methods:}
To further demonstrate the superiority of the effectiveness and generalizability of our method, we compare the state-of-the-art approaches in Tab.~\ref{general}. From the table, we have two following observations:
Firstly, all the state-of-the-art approaches~\cite{singh2022cass,zhang2022dino,VST,zhuge2022salient,zhao2020suppress,DIS,jia2022segment,COD} show less-desirable performances in each dataset. Instead, our BA-SAM (specialized models) consistently outperforms these methods when fine-tuned on each downstream dataset. Secondly, almost all of these state-of-the-art techniques are specifically designed for one task and cannot be generalized well to other tasks. Due to the strong zero-shot generalization capability of SAM~\cite{SAM}, our proposed BA-SAM can also be employed as a generalized model, which fine-tunes with all of these downstream datasets in a unified and shared model. Importantly, unlike~\cite{VST,singh2022cass,zhang2022dino,codella2018skin}, we eliminate the need for employing additional techniques to further enhance the performance. As shown in Tab.~\ref{general}, our generalized model also consistently promotes the performance of SAM on all datasets, demonstrating its remarkable generalizability.

\subsection{Ablation Study and Analysis} \label{sec:ablation}

In this section, we first conduct ablation studies to study
the contribution of each component. Then, we investigate
the impact of the new scaling factor (NSF) and bias-mode attention mask (BM-AM) with a more detailed analysis. 
% We conduct ablation studies on our proposed BA-SAM using MobileSAM as the backbone, analyzing the impact of the proposed Scaling Factor and Bias-Mode Attention Mask on the segmentation accuracy. We use four datasets from different tasks for ablation experiments, namely DIS5k, DUTS, ISIC, COD10K.

\noindent\textbf{Ablations Studies of Each Component.}
Tab.~\ref{tab:ablation} summarizes the effect of each designed component on the settings with and without fine-tuning, respectively. The baseline means using the MobileSAM~\cite{mobile_sam} as the base network that uses the vanilla scaling factor (VSF) in the attention layer~\cite{Vaswani2017Attention}. New Scaling Factor and Bias-Mode Attention Mask are abbreviated as NSF and BM-AM, respectively.
From the table, we observe that NSF could achieve superior performances than the baseline with VSF. This is because the vanilla attention in SAM~\cite{SAM} and MobileSAM~\cite{mobile_sam} does not consider maintaining the magnitude consistency when Softmax outputs change drastically due to input resolutions vary during the testing. In contrast, our NSF explicitly maintains the magnitude consistency and alleviates the performance degradation. Furthermore, by adding BM-AM, the performance could be further boosted when extrapolating to a larger test length. 
These improvements confirm that these individual components are complementary
and together they significantly promote the performance.

\begin{table}
    \centering
   \footnotesize
   \renewcommand{\arraystretch}{1.0}
   \setlength\tabcolsep{2.5pt}
    \begin{tabular}{r|cccc}
    \toprule[0.2em]
        Methods & ISIC~\cite{codella2018skin} & DUTS~\cite{DUTS} &  DIS-TE4~\cite{DIS} &   COD10K~\cite{COD}\\
    \midrule[0.2em]
 \multicolumn{5}{c}{Specialized models}\\
 \hline
         CASS~\cite{singh2022cass} & 0.086 & -&  - &   - \\
         DINO~\cite{zhang2022dino} & 0.081 & -& -  &   - \\
        MSA~\cite{wu2023medical} & 0.049 & -& -  &   - \\
         VST$_{21}$~\cite{VST} & -  & 0.037 & -  &  - \\ 
         ICONet$_{22}$~\cite{zhuge2022salient}& -  &0.037 & -  &  - \\
         Gate$_{20}$~\cite{zhao2020suppress}& -  & -&  0.109&  - \\
         IS-Net~\cite{DIS}&  -  &-&  0.072&   - \\
         % HQ-SAM~\cite{ke2023segment}&  -  &-&  \textbf{0.053}&   - \\
        SINet~\cite{COD} &-  & -  & -  & 0.092\\
        SegMaR$_{22}$~\cite{jia2022segment} & -  & -  & -  & 0.034\\
       % SAM-Adapter~\cite{chen2023sam} &-  & -  & -  & 0.026\\
    \hline
         \multicolumn{5}{c}{The same framework, 4 Specialized models}\\
    \hline
 \rowcolor{gray!15} Ours (BA-SAM) &  \textbf{0.053} & \textbf{0.031} & \textbf{0.055} & \textbf{0.029} \\
 \hline
 \multicolumn{5}{c}{Generalized model}\\
 \hline
 SAM~\cite{SAM}& 0.419 & 0.298& 0.373 &0.217 \\
\rowcolor{gray!15} Ours (BA-SAM)  & \textbf{0.054}  & \textbf{0.030} &\textbf{0.054}  & \textbf{0.054}\\
\bottomrule[0.2em]
    \end{tabular}
    % \vspace{-3mm}
    \caption{Comparison results (MAE) with state-of-the-art specialized models on various segmentation tasks. }
    \label{general}
\end{table}

\begin{table}[t]
  \centering
    \footnotesize
    \renewcommand{\arraystretch}{1.0}
    \setlength\tabcolsep{5pt}
  \begin{tabular}{r| c c c | c c c }
    \toprule[0.2em]
    Model &  AP & $AP_{50}$ & $AP_{75}$ & $AP_S$ & $AP_M$ & $AP_L$ \\
    \toprule[0.2em]
    SAM~\cite{SAM} &  42.5 & 69.6 & 44.7 & 29.7 & 47.0 & 56.7 \\
    \rowcolor{gray!15} Ours (w~\cite{SAM}) & 43.0 & 70.0 & 45.4 & 30.0 & 47.4 & 57.1 \\  
    \hline 
    MobileSAM~\cite{mobile_sam} &  40.8 & 68.4 & 41.6 & 26.0 & 44.4 & 57.6 \\  
    \rowcolor{gray!15} Ours (w~\cite{mobile_sam}) & 41.2 & 69.0 & 42.1 & 26.2 & 44.8 & 58.2 \\
    \bottomrule[0.2em]
  \end{tabular}
  % \vspace{-3mm}
   \caption{\small Results (AP) on COCO~\cite{lin2014microsoft} instance segmentation.}
   \label{coco}
  % \vspace{-4mm}
\end{table}

\noindent\textbf{Impact of Slope in Bias-Mode Attention Mask.}
\yr{In the Bias-Mode Attention Mask, the magnitude of the slope $\beta$ determines penalty rates in different heads.  
We found that the best performance is achieved when $\beta=0.1$. Besides,
our method is robust to different slope choices. In the zero-shot case, we use a fixed slope $\beta = 1$ by default. (More details could be referred to in the supplementary material)}

\noindent\textbf{Computational Efficiency.}\label{efficiency}
In Tab.~\ref{Computational Efficiency}, we analyze the computational efficiency between the baselines and our BA-SAM. 
All the experiments are conducted on the same NVIDIA RTX 4090GPU to ensure fair comparisons.
From the table, we observe that our BA-SAM is highly lightweight, incurring negligible computational overhead for the models. The reasons lie in two aspects: firstly, the NSF exhibits nearly identical computational complexity with the vanilla one. Besides, the BM-AM is seamlessly incorporated by adding a mask matrix to the query-key dot product before applying the Softmax operation. Although there is a slight increase in memory usage, it remains negligible compared to the memory occupied by large models.

\begin{table}[t!]
    \centering
\footnotesize
\renewcommand{\arraystretch}{1.0}
\setlength\tabcolsep{3.5pt}
    \begin{tabular}{l|cccc}
    \toprule[0.2em]
        Methods & ISIC~\cite{codella2018skin} & DUTS~\cite{DUTS} &  DIS-TE4~\cite{DIS} &   COD10K~\cite{COD}\\
    \midrule[0.2em]
 \multicolumn{5}{c}{Without fine-tuning}\\
 \hline
 Baseline~\cite{mobile_sam} & 17.8  & 6.5 & 11.7 & 11.9 \\
 + NSF  &  16.4 & 7.5 &11.3  & 12.0\\
  + BM-AM  & 16.8 & \textbf{7.2} & 11.7 & 11.9\\
  + Both  & \textbf{15.9}  & 7.9 &\textbf{10.9}  & \textbf{11.9}\\
    \hline
\multicolumn{5}{c}{With fine-tuning (1024)}\\
    \hline
 Baseline~\cite{mobile_sam}& 42.2  & 4.8 & 0.8 & 3.3 \\
 + NSF  & 40.9  & 4.6 &0.7  & 3.0\\
  + BM-AM  & 41.2 & 4.5 &0.8  & 2.7\\
  + Both  & \textbf{40.4}  & \textbf{4.4} & \textbf{0.6}  & \textbf{2.5} \\
 \hline
 \multicolumn{5}{c}{With fine-tuning (2048) }\\
 \hline
 Baseline~\cite{mobile_sam}& 14.4 & 4.6 & 1.0 &2.3 \\
 + NSF  &  \textbf{13.1} & 4.6 & 0.8 & 2.4\\
  + BM-AM  & 13.7  & 4.5 &0.9 & 2.3\\
  + Both  & 13.4  & \textbf{4.4} & \textbf{0.8} & \textbf{2.3}\\
\bottomrule[0.2em]
    \end{tabular}
    % \vspace{-3mm}
    \caption{Ablation study of each component on the settings with and without fine-tuning. Numbers indicate the performance degradation, $\Delta$\textit{diff}. A lower $\Delta$\textit{diff} means a better performance. 
    %($\Delta$\textit{diff} The lower the better.) 
    }
    \label{tab:ablation}
\end{table}

\begin{table}[t]
  \centering
\footnotesize
\renewcommand{\arraystretch}{1.0}
\setlength\tabcolsep{7pt}
  \begin{tabular}{r| c |c | c  }
    \toprule[0.2em]
    Model &  Params (M) & Speed (ms)& Train Hours (h) \\
    \toprule[0.2em]
    SAM~\cite{SAM} &  81 &  113.9 & -  \\
    \rowcolor{gray!15} Ours (w~\cite{SAM}) & 81 & 114.0 & -  \\  
    \hline 
    MobileSAM~\cite{mobile_sam} &  9.66 & 16.2 & 0.64  \\  
    % HQ-SAM & 40.3M & 41.6 & 69.0 & 42.3 & 26.2 & 44.8 & 58.4 \\
    \rowcolor{gray!15} Ours (w~\cite{mobile_sam}) & 9.67 & 16.5 & 0.65 \\
    % $ Ours_{j} $ & 38.6M &  &  &  &  &  &  \\    
    \bottomrule[0.2em]
  \end{tabular}
  % \vspace{-1mm}
   \caption{\small Comparisons of computational efficiency between the baselines and our BA-SAM. Params: number of parameters. Speed: inference speed. The top part is performed on the zero-shot setting, and the bottom part is the scenarios with fine-tuning.
   %in the  RTX 4090 GPU.
   }
   \label{Computational Efficiency}
  % \vspace{-1mm}
\end{table}
\section{Conclusion}

In this paper, we address the important problem of varying image resolutions in SAM models by reformulating it as a problem of length extrapolation. To enhance the length extrapolation capability of SAM, we propose the Scalable Bias-mode Attention Mask (BA-SAM) SAM. A new scaling factor (NSF) is introduced to maintain the consistent magnitude of attention. In addition, a bias-mode attention mask (BM-AM) is designed to prioritize neighboring information, mitigating the impact of untrained distant information. 
Extensive evaluation of diverse datasets reveals its ability to significantly alleviate performance degradation in the zero-shot setting and achieve state-of-the-art performance with minimal fine-tuning.  Furthermore, we proposed a generalized model and benchmark, showcasing BA-SAM's generalizability across all four datasets. 

\section{Acknowledgement}
Shanghai Municipal Science and Technology Major Project (2021SHZDZX0102), Shanghai Science and Technology Commission  (21511101200), National Natural Science Foundation of China (No. 72192821), YuCaiKe [2023] Project Number: 14105167-2023

{
    \small
    \bibliographystyle{ieeenat_fullname}
    \bibliography{main}

\begin{thebibliography}{96}
\providecommand{\natexlab}[1]{#1}
\providecommand{\url}[1]{\texttt{#1}}
\expandafter\ifx\csname urlstyle\endcsname\relax
  \providecommand{\doi}[1]{doi: #1}\else
  \providecommand{\doi}{doi: \begingroup \urlstyle{rm}\Url}\fi

\bibitem[Bahdanau et~al.(2014)Bahdanau, Cho, and Bengio]{bahdanau2014neural}
Dzmitry Bahdanau, Kyunghyun Cho, and Yoshua Bengio.
\newblock Neural machine translation by jointly learning to align and translate.
\newblock \emph{arXiv preprint arXiv:1409.0473}, 2014.

\bibitem[Bello(2021)]{45}
Irwan Bello.
\newblock Lambdanetworks: Modeling long-range interactions without attention.
\newblock In \emph{CVPR}, 2021.

\bibitem[Bello et~al.(2019)Bello, Zoph, Le, Vaswani, and Shlens]{38}
Irwan Bello, Barret Zoph, Quoc Le, Ashish Vaswani, and Jonathon Shlens.
\newblock Attention augmented convolutional networks.
\newblock In \emph{ICCV}, 2019.

\bibitem[Beyer et~al.(2023)Beyer, Izmailov, Kolesnikov, Caron, Kornblith, Zhai, Minderer, Tschannen, Alabdulmohsin, and Pavetic]{beyer2023flexivit}
Lucas Beyer, Pavel Izmailov, Alexander Kolesnikov, Mathilde Caron, Simon Kornblith, Xiaohua Zhai, Matthias Minderer, Michael Tschannen, Ibrahim Alabdulmohsin, and Filip Pavetic.
\newblock Flexivit: One model for all patch sizes.
\newblock In \emph{CVPR}, 2023.

\bibitem[Bommasani et~al.(2021)Bommasani, Hudson, Adeli, Altman, Arora, von Arx, Bernstein, Bohg, Bosselut, Brunskill, et~al.]{bommasani2021opportunities}
Rishi Bommasani, Drew~A Hudson, Ehsan Adeli, Russ Altman, Simran Arora, Sydney von Arx, Michael~S Bernstein, Jeannette Bohg, Antoine Bosselut, Emma Brunskill, et~al.
\newblock On the opportunities and risks of foundation models.
\newblock \emph{arXiv preprint arXiv:2108.07258}, 2021.

\bibitem[Cai et~al.(2020)Cai, Gan, Wang, Zhang, and Han]{8}
Han Cai, Chuang Gan, Tianzhe Wang, Zhekai Zhang, and Song Han.
\newblock Once for all: Train one network and specialize it for efficient deployment.
\newblock \emph{ICLR}, 2020.

\bibitem[Chen et~al.(2022)Chen, Ge, Tong, Wang, Song, Wang, and Luo]{chen2022adaptformer}
Shoufa Chen, Chongjian Ge, Zhan Tong, Jiangliu Wang, Yibing Song, Jue Wang, and Ping Luo.
\newblock Adaptformer: Adapting vision transformers for scalable visual recognition.
\newblock \emph{NeurIPS}, 2022.

\bibitem[Chen et~al.(2023)Chen, Zhu, Ding, Cao, Zhang, Wang, Li, Sun, Mao, and Zang]{chen2023sam}
Tianrun Chen, Lanyun Zhu, Chaotao Ding, Runlong Cao, Shangzhan Zhang, Yan Wang, Zejian Li, Lingyun Sun, Papa Mao, and Ying Zang.
\newblock Sam fails to segment anything?--sam-adapter: Adapting sam in underperformed scenes: Camouflage, shadow, and more.
\newblock In \emph{arXiv preprint arXiv:2304.09148}, 2023.

\bibitem[Chen and He(2021)]{chen2021exploring}
Xinlei Chen and Kaiming He.
\newblock Exploring simple siamese representation learning.
\newblock In \emph{CVPR}, 2021.

\bibitem[Chen et~al.(2020)Chen, Dai, Liu, Chen, Yuan, and Liu]{40}
Yinpeng Chen, Xiyang Dai, Mengchen Liu, Dongdong Chen, Lu Yuan, and Zicheng Liu.
\newblock Dynamic convolution: Attention over convolution kernels.
\newblock In \emph{CVPR}, 2020.

\bibitem[Chi et~al.(2022)Chi, Fan, Ramadge, and Rudnicky]{chi2022kerple}
Ta-Chung Chi, Ting-Han Fan, Peter~J Ramadge, and Alexander Rudnicky.
\newblock Kerple: Kernelized relative positional embedding for length extrapolation.
\newblock \emph{NeurIPS}, 35, 2022.

\bibitem[Codella et~al.(2018)Codella, Gutman, Celebi, Helba, Marchetti, Dusza, Kalloo, Liopyris, Mishra, Kittler, et~al.]{codella2018skin}
Noel~CF Codella, David Gutman, M~Emre Celebi, Brian Helba, Michael~A Marchetti, Stephen~W Dusza, Aadi Kalloo, Konstantinos Liopyris, Nabin Mishra, Harald Kittler, et~al.
\newblock Skin lesion analysis toward melanoma detection: A challenge at the 2017 international symposium on biomedical imaging (isbi), hosted by the international skin imaging collaboration (isic).
\newblock In \emph{ISBI}, 2018.

\bibitem[Dehghani et~al.(2023)Dehghani, Mustafa, Djolonga, Heek, Minderer, Caron, Steiner, Puigcerver, Geirhos, Alabdulmohsin, et~al.]{dehghani2023patch}
Mostafa Dehghani, Basil Mustafa, Josip Djolonga, Jonathan Heek, Matthias Minderer, Mathilde Caron, Andreas Steiner, Joan Puigcerver, Robert Geirhos, Ibrahim Alabdulmohsin, et~al.
\newblock Patch n'pack: Navit, a vision transformer for any aspect ratio and resolution.
\newblock \emph{arXiv preprint arXiv:2307.06304}, 2023.

\bibitem[Dosovitskiy et~al.(2020)Dosovitskiy, Beyer, Kolesnikov, Weissenborn, Zhai, Unterthiner, Dehghani, Minderer, Heigold, Gelly, et~al.]{dosovitskiy2020image}
Alexey Dosovitskiy, Lucas Beyer, Alexander Kolesnikov, Dirk Weissenborn, Xiaohua Zhai, Thomas Unterthiner, Mostafa Dehghani, Matthias Minderer, Georg Heigold, Sylvain Gelly, et~al.
\newblock An image is worth 16x16 words: Transformers for image recognition at scale.
\newblock \emph{arXiv preprint arXiv:2010.11929}, 2020.

\bibitem[Fan et~al.(2018)Fan, Cheng, Liu, Gao, Hou, and Borji]{fan2018salient}
Deng-Ping Fan, Ming-Ming Cheng, Jiang-Jiang Liu, Shang-Hua Gao, Qibin Hou, and Ali Borji.
\newblock Salient objects in clutter: Bringing salient object detection to the foreground.
\newblock In \emph{ECCV}, 2018.

\bibitem[Fan et~al.(2020)Fan, Ji, Sun, Cheng, Shen, and Shao]{COD}
Deng-Ping Fan, Ge-Peng Ji, Guolei Sun, Ming-Ming Cheng, Jianbing Shen, and Ling Shao.
\newblock Camouflaged object detection.
\newblock In \emph{CVPR}, 2020.

\bibitem[Fan et~al.(2021{\natexlab{a}})Fan, Ji, Cheng, and Shao]{fan2021concealed}
Deng-Ping Fan, Ge-Peng Ji, Ming-Ming Cheng, and Ling Shao.
\newblock Concealed object detection.
\newblock \emph{TPAMI}, 2021{\natexlab{a}}.

\bibitem[Fan et~al.(2021{\natexlab{b}})Fan, Lin, Zhang, Zhu, and Cheng]{SIP}
Deng-Ping Fan, Zheng Lin, Zhao Zhang, Menglong Zhu, and Ming-Ming Cheng.
\newblock Rethinking rgb-d salient object detection: Models, data sets, and large-scale benchmarks.
\newblock \emph{TNNLS}, 2021{\natexlab{b}}.

\bibitem[Fan et~al.(2022{\natexlab{a}})Fan, Li, Lin, Ji, Zhang, Cheng, Fu, and Shen]{fan2022rething}
Deng-Ping Fan, Tengpeng Li, Zheng Lin, Ge-Peng Ji, Dingwen Zhang, Ming-Ming Cheng, Huazhu Fu, and Jianbing Shen.
\newblock Re-thinking co-salient object detection.
\newblock \emph{TPAMI}, 2022{\natexlab{a}}.

\bibitem[Fan et~al.(2022{\natexlab{b}})Fan, Zhang, Xu, Cheng, and Shao]{fan2022salient}
Deng-Ping Fan, Jing Zhang, Gang Xu, Ming-Ming Cheng, and Ling Shao.
\newblock Salient objects in clutter.
\newblock \emph{TPAMI}, 2022{\natexlab{b}}.

\bibitem[Fang et~al.(2023)Fang, Wang, Xie, Sun, Wu, Wang, Huang, Wang, and Cao]{fang2023eva}
Yuxin Fang, Wen Wang, Binhui Xie, Quan Sun, Ledell Wu, Xinggang Wang, Tiejun Huang, Xinlong Wang, and Yue Cao.
\newblock Eva: Exploring the limits of masked visual representation learning at scale.
\newblock In \emph{CVPR}, 2023.

\bibitem[Feng et~al.(2022)Feng, Zhou, Gu, Tan, Cheng, Lu, Shi, and Ma]{feng2022dmt}
Zhengyang Feng, Qianyu Zhou, Qiqi Gu, Xin Tan, Guangliang Cheng, Xuequan Lu, Jianping Shi, and Lizhuang Ma.
\newblock Dmt: Dynamic mutual training for semi-supervised learning.
\newblock \emph{Pattern Recognition}, 2022.

\bibitem[Gehring et~al.(2017)Gehring, Auli, Grangier, Yarats, and Dauphin]{gehring2017convolutional}
Jonas Gehring, Michael Auli, David Grangier, Denis Yarats, and Yann~N Dauphin.
\newblock Convolutional sequence to sequence learning.
\newblock In \emph{ICML}, 2017.

\bibitem[Gong et~al.(2021)Gong, Wang, Li, Chen, Yan, Tian, Chandra, et~al.]{18}
Chengyue Gong, Dilin Wang, Meng Li, Xinlei Chen, Zhicheng Yan, Yuandong Tian, Vikas Chandra, et~al.
\newblock Nasvit: Neural architecture search for efficient vision transformers with gradient conflict aware supernet training.
\newblock In \emph{ICLR}, 2021.

\bibitem[Graham et~al.(2021)Graham, El-Nouby, Touvron, Stock, Joulin, J{\'e}gou, and Douze]{graham2021levit}
Benjamin Graham, Alaaeldin El-Nouby, Hugo Touvron, Pierre Stock, Armand Joulin, Herv{\'e} J{\'e}gou, and Matthijs Douze.
\newblock Levit: a vision transformer in convnet's clothing for faster inference.
\newblock In \emph{ICCV}, 2021.

\bibitem[Gu et~al.(2021)Gu, Zhou, Xu, Feng, Cheng, Lu, Shi, and Ma]{gu2021pit}
Qiqi Gu, Qianyu Zhou, Minghao Xu, Zhengyang Feng, Guangliang Cheng, Xuequan Lu, Jianping Shi, and Lizhuang Ma.
\newblock Pit: Position-invariant transform for cross-fov domain adaptation.
\newblock In \emph{ICCV}, 2021.

\bibitem[He et~al.(2021{\natexlab{a}})He, Zhou, Ma, Berg-Kirkpatrick, and Neubig]{He_Zhou_Ma_Berg-Kirkpatrick_Neubig_2021}
Junxian He, Chunting Zhou, Xuezhe Ma, Taylor Berg-Kirkpatrick, and Graham Neubig.
\newblock Towards a unified view of parameter-efficient transfer learning.
\newblock In \emph{ICLR}, 2021{\natexlab{a}}.

\bibitem[He et~al.(2016)He, Zhang, Ren, and Sun]{he2016deep}
Kaiming He, Xiangyu Zhang, Shaoqing Ren, and Jian Sun.
\newblock Deep residual learning for image recognition.
\newblock In \emph{CVPR}, 2016.

\bibitem[He et~al.(2022)He, Chen, Xie, Li, Dollar, and Girshick]{20}
Kaiming He, Xinlei Chen, Saining Xie, Yanghao Li, Piotr Dollar, and Ross Girshick.
\newblock Masked autoencoders are scalable vision learners.
\newblock In \emph{CVPR}, 2022.

\bibitem[He et~al.(2021{\natexlab{b}})He, Zhou, Li, Niu, Cheng, Li, Liu, Tong, Ma, and Zhang]{he2021end}
Lu He, Qianyu Zhou, Xiangtai Li, Li Niu, Guangliang Cheng, Xiao Li, Wenxuan Liu, Yunhai Tong, Lizhuang Ma, and Liqing Zhang.
\newblock End-to-end video object detection with spatial-temporal transformers.
\newblock In \emph{ACM MM}, 2021{\natexlab{b}}.

\bibitem[Ji et~al.(2023{\natexlab{a}})Ji, Fan, Xu, Cheng, Zhou, and Van~Gool]{ji2023sam}
Ge-Peng Ji, Deng-Ping Fan, Peng Xu, Ming-Ming Cheng, Bowen Zhou, and Luc Van~Gool.
\newblock Sam struggles in concealed scenes--empirical study on" segment anything".
\newblock \emph{SCIS}, 2023{\natexlab{a}}.

\bibitem[Ji et~al.(2023{\natexlab{b}})Ji, Li, Bi, Li, and Cheng]{ji2023segment}
Wei Ji, Jingjing Li, Qi Bi, Wenbo Li, and Li Cheng.
\newblock Segment anything is not always perfect: An investigation of sam on different real-world applications.
\newblock \emph{arXiv preprint arXiv:2304.05750}, 2023{\natexlab{b}}.

\bibitem[Jia et~al.(2021)Jia, Yang, Xia, Chen, Parekh, Pham, Le, Sung, Li, and Duerig]{align_icml}
Chao Jia, Yinfei Yang, Ye Xia, Yi-Ting Chen, Zarana Parekh, Hieu Pham, Quoc Le, Yun-Hsuan Sung, Zhen Li, and Tom Duerig.
\newblock Scaling up visual and vision-language representation learning with noisy text supervision.
\newblock In \emph{ICML}, 2021.

\bibitem[Jia et~al.(2022{\natexlab{a}})Jia, Tang, Chen, Cardie, Belongie, Hariharan, and Lim]{jia2022visual}
Menglin Jia, Luming Tang, Bor-Chun Chen, Claire Cardie, Serge Belongie, Bharath Hariharan, and Ser-Nam Lim.
\newblock Visual prompt tuning.
\newblock In \emph{ECCV}, 2022{\natexlab{a}}.

\bibitem[Jia et~al.(2022{\natexlab{b}})Jia, Yao, Liu, Fan, Liu, and Luo]{jia2022segment}
Qi Jia, Shuilian Yao, Yu Liu, Xin Fan, Risheng Liu, and Zhongxuan Luo.
\newblock Segment, magnify and reiterate: Detecting camouflaged objects the hard way.
\newblock In \emph{CVPR}, 2022{\natexlab{b}}.

\bibitem[Kirillov et~al.(2023)Kirillov, Mintun, Ravi, Mao, Rolland, Gustafson, Xiao, Whitehead, Berg, Lo, et~al.]{SAM}
Alexander Kirillov, Eric Mintun, Nikhila Ravi, Hanzi Mao, Chloe Rolland, Laura Gustafson, Tete Xiao, Spencer Whitehead, Alexander~C Berg, Wan-Yen Lo, et~al.
\newblock Segment anything.
\newblock \emph{arXiv preprint arXiv:2304.02643}, 2023.

\bibitem[Kusupati et~al.(2022)Kusupati, Bhatt, Rege, Wallingford, Sinha, Ramanujan, Howard-Snyder, Chen, Kakade, Jain, and Farhadi]{31_felex}
Aditya Kusupati, Gantavya Bhatt, Aniket Rege, Matthew Wallingford, Aditya Sinha, Vivek Ramanujan, William Howard-Snyder, Kaifeng Chen, Sham Kakade, Prateek Jain, and Ali Farhadi.
\newblock Matryoshka representations for adaptive deployment.
\newblock \emph{arXiv preprint arXiv:2205.13147}, 2022.

\bibitem[Lee et~al.(2023)Lee, Joshi, Turc, Hu, Liu, Eisenschlos, Khandelwal, Shaw, Chang, and Toutanova]{lee2023pix2struct}
Kenton Lee, Mandar Joshi, Iulia~Raluca Turc, Hexiang Hu, Fangyu Liu, Julian~Martin Eisenschlos, Urvashi Khandelwal, Peter Shaw, Ming-Wei Chang, and Kristina Toutanova.
\newblock Pix2struct: Screenshot parsing as pretraining for visual language understanding.
\newblock In \emph{ICML}, 2023.

\bibitem[Li et~al.(2019)Li, Wang, Hu, and Yang]{37}
Xiang Li, Wenhai Wang, Xiaolin Hu, and Jian Yang.
\newblock Selective kernel networks.
\newblock In \emph{CVPR}, 2019.

\bibitem[Li et~al.(2023)Li, Ding, Zhang, Yuan, Cheng, Jiangmiao, Chen, Liu, and Loy]{li2023transformer}
Xiangtai Li, Henghui Ding, Wenwei Zhang, Haobo Yuan, Guangliang Cheng, Pang Jiangmiao, Kai Chen, Ziwei Liu, and Chen~Change Loy.
\newblock Transformer-based visual segmentation: A survey.
\newblock \emph{arXiv pre-print}, 2023.

\bibitem[Li et~al.(2024)Li, Yuan, Li, Ding, Wu, Zhang, Li, Chen, and Loy]{li2024omg}
Xiangtai Li, Haobo Yuan, Wei Li, Henghui Ding, Size Wu, Wenwei Zhang, Yining Li, Kai Chen, and Chen~Change Loy.
\newblock Omg-seg: Is one model good enough for all segmentation?
\newblock \emph{CVPR}, 2024.

\bibitem[Li and Liang(2021)]{Li_Liang_2021}
Xiang~Lisa Li and Percy Liang.
\newblock Prefix-tuning: Optimizing continuous prompts for generation.
\newblock In \emph{ACL)}, 2021.

\bibitem[Liang et~al.(2022)Liang, Zadeh, and Morency]{liang2022foundations}
Paul~Pu Liang, Amir Zadeh, and Louis-Philippe Morency.
\newblock Foundations and recent trends in multimodal machine learning: Principles, challenges, and open questions.
\newblock \emph{arXiv preprint arXiv:2209.03430}, 2022.

\bibitem[Lin et~al.(2022)Lin, Chen, Zhang, Li, Shen, Shen, and Ji]{34_flex}
Mingbao Lin, Mengzhao Chen, Yuxin Zhang, Ke Li, Yunhang Shen, Chunhua Shen, and Rongrong Ji.
\newblock Super vision transformer.
\newblock \emph{IJCV}, 2022.

\bibitem[Lin et~al.(2014)Lin, Maire, Belongie, Hays, Perona, Ramanan, Doll{\'a}r, and Zitnick]{lin2014microsoft}
Tsung-Yi Lin, Michael Maire, Serge Belongie, James Hays, Pietro Perona, Deva Ramanan, Piotr Doll{\'a}r, and C~Lawrence Zitnick.
\newblock Microsoft coco: Common objects in context.
\newblock In \emph{ECCV}, 2014.

\bibitem[Liu et~al.(2021{\natexlab{a}})Liu, Zhang, Wan, Shao, and Han]{VST}
Nian Liu, Ni Zhang, Kaiyuan Wan, Ling Shao, and Junwei Han.
\newblock Visual saliency transformer.
\newblock In \emph{ICCV}, 2021{\natexlab{a}}.

\bibitem[Liu et~al.(2021{\natexlab{b}})Liu, Lin, Cao, Hu, Wei, Zhang, Lin, and Guo]{liu2021swin}
Ze Liu, Yutong Lin, Yue Cao, Han Hu, Yixuan Wei, Zheng Zhang, Stephen Lin, and Baining Guo.
\newblock Swin transformer: Hierarchical vision transformer using shifted windows.
\newblock In \emph{ICCV}, 2021{\natexlab{b}}.

\bibitem[Liu et~al.(2022)Liu, Hu, Lin, Yao, Xie, Wei, Ning, Cao, Zhang, Dong, et~al.]{liu2022swin}
Ze Liu, Han Hu, Yutong Lin, Zhuliang Yao, Zhenda Xie, Yixuan Wei, Jia Ning, Yue Cao, Zheng Zhang, Li Dong, et~al.
\newblock Swin transformer v2: Scaling up capacity and resolution.
\newblock In \emph{CVPR}, 2022.

\bibitem[Long et~al.(2023{\natexlab{a}})Long, Zhou, Ying, Ma, and Luo]{long2023diverse}
Shaocong Long, Qianyu Zhou, Chenhao Ying, Lizhuang Ma, and Yuan Luo.
\newblock Diverse target and contribution scheduling for domain generalization.
\newblock \emph{arXiv preprint arXiv:2309.16460}, 2023{\natexlab{a}}.

\bibitem[Long et~al.(2023{\natexlab{b}})Long, Zhou, Ying, Ma, and Luo]{long2023rethink}
Shaocong Long, Qianyu Zhou, Chenhao Ying, Lizhuang Ma, and Yuan Luo.
\newblock Rethinking domain generalization: Discriminability and generalizability.
\newblock \emph{arXiv preprint arXiv:2309.16483}, 2023{\natexlab{b}}.

\bibitem[Parmar et~al.(2018)Parmar, Vaswani, Uszkoreit, Kaiser, Shazeer, Ku, and Tran]{parmar2018image}
Niki Parmar, Ashish Vaswani, Jakob Uszkoreit, Lukasz Kaiser, Noam Shazeer, Alexander Ku, and Dustin Tran.
\newblock Image transformer.
\newblock In \emph{ICML}, 2018.

\bibitem[Press et~al.(2021)Press, Smith, Lewis, et~al.]{press2021train}
Ofir Press, Noah~A Smith, Mike Lewis, et~al.
\newblock Train short, test long: Attention with linear biases enables input length extrapolation.
\newblock \emph{arXiv preprint arXiv:2108.12409}, 2021.

\bibitem[Qin et~al.(2022)Qin, Dai, Hu, Fan, Shao, and Van~Gool]{DIS}
Xuebin Qin, Hang Dai, Xiaobin Hu, Deng-Ping Fan, Ling Shao, and Luc Van~Gool.
\newblock Highly accurate dichotomous image segmentation.
\newblock In \emph{ECCV}, 2022.

\bibitem[Radford et~al.(2021)Radford, Kim, Hallacy, Ramesh, Goh, Agarwal, Sastry, Askell, Mishkin, Clark, et~al.]{CLIP}
Alec Radford, Jong~Wook Kim, Chris Hallacy, Aditya Ramesh, Gabriel Goh, Sandhini Agarwal, Girish Sastry, Amanda Askell, Pamela Mishkin, Jack Clark, et~al.
\newblock Learning transferable visual models from natural language supervision.
\newblock In \emph{ICML}, 2021.

\bibitem[Ramesh et~al.(2022)Ramesh, Dhariwal, Nichol, Chu, and Chen]{ramesh2022hierarchical}
Aditya Ramesh, Prafulla Dhariwal, Alex Nichol, Casey Chu, and Mark Chen.
\newblock Hierarchical text-conditional image generation with clip latents.
\newblock \emph{arXiv preprint arXiv:2204.06125}, 2022.

\bibitem[Shaw et~al.(2018)Shaw, Uszkoreit, and Vaswani]{shaw2018self}
Peter Shaw, Jakob Uszkoreit, and Ashish Vaswani.
\newblock Self-attention with relative position representations.
\newblock In \emph{Proceedings of NAACL-HLT}, 2018.

\bibitem[Singh(2022)]{singh2022cass}
et~al Singh, Pranav.
\newblock Cass: cross architectural self-supervision for medical image analysis.
\newblock \emph{arXiv preprint arXiv:2206.04170}, 2022.

\bibitem[Song et~al.(2023)Song, Zhou, and Ma]{song2023rethinking}
Yiran Song, Qianyu Zhou, and Lizhuang Ma.
\newblock Rethinking implicit neural representations for vision learners.
\newblock In \emph{ICASSP}, 2023.

\bibitem[Srinivas et~al.(2021)Srinivas, Lin, Parmar, Shlens, Abbeel, and Vaswani]{46}
Aravind Srinivas, Tsung-Yi Lin, Niki Parmar, Jonathon Shlens, Pieter Abbeel, and Ashish Vaswani.
\newblock Bottleneck transformers for visual recognition.
\newblock In \emph{CVPR}, 2021.

\bibitem[Su(2023)]{kexuefm}
JianLin Su.
\newblock Transformer upgrade road: 7, length extrapolation and local attention, 2023.

\bibitem[Szegedy et~al.(2015)Szegedy, Liu, Jia, Sermanet, Reed, Anguelov, Erhan, Vanhoucke, and Rabinovich]{42}
Christian Szegedy, Wei Liu, Yangqing Jia, Pierre Sermanet, Scott Reed, Dragomir Anguelov, Dumitru Erhan, Vincent Vanhoucke, and Andrew Rabinovich.
\newblock Going deeper with convolutions.
\newblock In \emph{CVPR}, 2015.

\bibitem[Vaswani et~al.(2017)Vaswani, Shazeer, Parmar, Uszkoreit, Jones, Gomez, Kaiser, and Polosukhin]{Vaswani2017Attention}
Ashish Vaswani, Noam Shazeer, Niki Parmar, Jakob Uszkoreit, Llion Jones, Aidan~N. Gomez, Lukasz Kaiser, and Illia Polosukhin.
\newblock Attention is all you need.
\newblock In \emph{{NeurIPS}}, 2017.

\bibitem[Wang et~al.(2017{\natexlab{a}})Wang, Jiang, Qian, Yang, Li, Zhang, Wang, and Tang]{35}
Fei Wang, Mengqing Jiang, Chen Qian, Shuo Yang, Cheng Li, Honggang Zhang, Xiaogang Wang, and Xiaoou Tang.
\newblock Residual attention network for image classification.
\newblock In \emph{CVPR}, 2017{\natexlab{a}}.

\bibitem[Wang et~al.(2017{\natexlab{b}})Wang, Lu, Wang, Feng, Wang, Yin, and Ruan]{DUTS}
Lijun Wang, Huchuan Lu, Yifan Wang, Mengyang Feng, Dong Wang, Baocai Yin, and Xiang Ruan.
\newblock Learning to detect salient objects with image-level supervision.
\newblock In \emph{CVPR}, 2017{\natexlab{b}}.

\bibitem[Wang et~al.(2021)Wang, Xie, Li, Fan, Song, Liang, Lu, Luo, and Shao]{48}
Wenhai Wang, Enze Xie, Xiang Li, Deng-Ping Fan, Kaitao Song, Ding Liang, Tong Lu, Ping Luo, and Ling Shao.
\newblock Pyramid vision transformer: A versatile backbone for dense prediction without convolutions.
\newblock In \emph{ICCV}, 2021.

\bibitem[Wang et~al.(2018)Wang, Girshick, Gupta, and He]{39}
Xiaolong Wang, Ross Girshick, Abhinav Gupta, and Kaiming He.
\newblock Non-local neural networks.
\newblock In \emph{CVPR}, 2018.

\bibitem[Wang et~al.(2023)Wang, Chen, Qian, Gao, Wei, Wang, Tian, and Gao]{wang2023large}
Xiao Wang, Guangyao Chen, Guangwu Qian, Pengcheng Gao, Xiao-Yong Wei, Yaowei Wang, Yonghong Tian, and Wen Gao.
\newblock Large-scale multi-modal pre-trained models: A comprehensive survey.
\newblock \emph{MIR}, 2023.

\bibitem[Woo et~al.(2018)Woo, Park, Lee, and Kweon]{36}
Sanghyun Woo, Jongchan Park, Joon-Young Lee, and In~So Kweon.
\newblock Cbam: Convolutional block attention module.
\newblock In \emph{ECCV}, 2018.

\bibitem[Wu et~al.(2020{\natexlab{a}})Wu, Xu, Dai, Wan, Zhang, Yan, Tomizuka, Gonzalez, Keutzer, and Vajda]{23}
Bichen Wu, Chenfeng Xu, Xiaoliang Dai, Alvin Wan, Peizhao Zhang, Zhicheng Yan, Masayoshi Tomizuka, Joseph Gonzalez, Kurt Keutzer, and Peter Vajda.
\newblock Generating long sequences with sparse transformers.
\newblock \emph{arXiv preprint arXiv:2006.03677}, 2020{\natexlab{a}}.

\bibitem[Wu et~al.(2020{\natexlab{b}})Wu, Xu, Dai, Wan, Zhang, Yan, Tomizuka, Gonzalez, Keutzer, and Vajda]{47}
Bichen Wu, Chenfeng Xu, Xiaoliang Dai, Alvin Wan, Peizhao Zhang, Zhicheng Yan, Masayoshi Tomizuka, Joseph Gonzalez, Kurt Keutzer, and Peter Vajda.
\newblock Visual transformers: Token-based image representation and processing for computer vision.
\newblock \emph{arXiv preprint arXiv:2006.03677}, 2020{\natexlab{b}}.

\bibitem[Wu et~al.(2023)Wu, Fu, Fang, Liu, Wang, Xu, Jin, and Arbel]{wu2023medical}
Junde Wu, Rao Fu, Huihui Fang, Yuanpei Liu, Zhaowei Wang, Yanwu Xu, Yueming Jin, and Tal Arbel.
\newblock Medical sam adapter: Adapting segment anything model for medical image segmentation.
\newblock \emph{arXiv preprint arXiv:2304.12620}, 2023.

\bibitem[Wu et~al.(2024)Wu, Li, Xu, Yuan, Ding, Yang, Li, Zhang, Tong, Jiang, Ghanem, and Tao]{wu2023open}
Jianzong Wu, Xiangtai Li, Shilin Xu, Haobo Yuan, Henghui Ding, Yibo Yang, Xia Li, Jiangning Zhang, Yunhai Tong, Xudong Jiang, Bernard Ghanem, and Dacheng Tao.
\newblock Towards open vocabulary learning: A survey.
\newblock \emph{T-PAMI}, 2024.

\bibitem[Wu et~al.(2022)Wu, Zhang, Peng, Liu, Xiao, Fu, and Yuan]{wu2022tinyvit}
Kan Wu, Jinnian Zhang, Houwen Peng, Mengchen Liu, Bin Xiao, Jianlong Fu, and Lu Yuan.
\newblock Tinyvit: Fast pretraining distillation for small vision transformers.
\newblock In \emph{ECCV}, 2022.

\bibitem[Xie et~al.(2022)Xie, Zhang, Cao, Lin, Bao, Yao, Dai, and Hu]{xie2022simmim}
Zhenda Xie, Zheng Zhang, Yue Cao, Yutong Lin, Jianmin Bao, Zhuliang Yao, Qi Dai, and Han Hu.
\newblock Simmim: A simple framework for masked image modeling.
\newblock In \emph{CVPR}, 2022.

\bibitem[Xu et~al.(2021)Xu, Liu, Zhou, Hao, Cao, Feng, and Ma]{xu2021semi}
Hongyi Xu, Fengqi Liu, Qianyu Zhou, Jinkun Hao, Zhijie Cao, Zhengyang Feng, and Lizhuang Ma.
\newblock Semi-supervised 3d object detection via adaptive pseudo-labeling.
\newblock In \emph{ICIP}, 2021.

\bibitem[Yu et~al.(2020)Yu, Jin, Liu, Bender, Kindermans, Tan, Huang, Song, Pang, and Le]{63}
Jiahui Yu, Pengchong Jin, Hanxiao Liu, Gabriel Bender, Pieter-Jan Kindermans, Mingxing Tan, Thomas Huang, Xiaodan Song, Ruoming Pang, and Quoc Le.
\newblock Bignas: Scaling up neural architecture search with big single-stage models.
\newblock In \emph{ECCV}, 2020.

\bibitem[Yuan et~al.(2024)Yuan, Li, Zhou, Li, Chen, and Loy]{yuan2024ovsam}
Haobo Yuan, Xiangtai Li, Chong Zhou, Yining Li, Kai Chen, and Chen~Change Loy.
\newblock Open-vocabulary sam: Segment and recognize twenty-thousand classes interactively.
\newblock \emph{arXiv preprint}, 2024.

\bibitem[Zhang et~al.(2023)Zhang, Han, Qiao, Kim, Bae, Lee, and Hong]{mobile_sam}
Chaoning Zhang, Dongshen Han, Yu Qiao, Jung~Uk Kim, Sung-Ho Bae, Seungkyu Lee, and Choong~Seon Hong.
\newblock Faster segment anything: Towards lightweight sam for mobile applications.
\newblock \emph{arXiv preprint arXiv:2306.14289}, 2023.

\bibitem[Zhang et~al.(2022{\natexlab{a}})Zhang, Li, Liu, Zhang, Su, Zhu, Ni, and Shum]{zhang2022dino}
Hao Zhang, Feng Li, Shilong Liu, Lei Zhang, Hang Su, Jun Zhu, Lionel~M Ni, and Heung-Yeung Shum.
\newblock Dino: Detr with improved denoising anchor boxes for end-to-end object detection.
\newblock \emph{arXiv preprint arXiv:2203.03605}, 2022{\natexlab{a}}.

\bibitem[Zhang et~al.(2022{\natexlab{b}})Zhang, Wu, Zhang, Zhu, Lin, Zhang, Sun, He, Mueller, Manmatha, et~al.]{43}
Hang Zhang, Chongruo Wu, Zhongyue Zhang, Yi Zhu, Haibin Lin, Zhi Zhang, Yue Sun, Tong He, Jonas Mueller, R Manmatha, et~al.
\newblock Resnest: Split-attention networks.
\newblock In \emph{CVPR}, 2022{\natexlab{b}}.

\bibitem[Zhang and Liu(2023)]{zhang2023customized}
Kaidong Zhang and Dong Liu.
\newblock Customized segment anything model for medical image segmentation.
\newblock In \emph{arXiv preprint arXiv:2304.13785}, 2023.

\bibitem[Zhao et~al.(2020{\natexlab{a}})Zhao, Jia, and Koltun]{41}
Hengshuang Zhao, Jiaya Jia, and Vladlen Koltun.
\newblock Exploring self-attention for image recognition.
\newblock In \emph{CVPR}, 2020{\natexlab{a}}.

\bibitem[Zhao et~al.(2020{\natexlab{b}})Zhao, Pang, Zhang, Lu, and Zhang]{zhao2020suppress}
Xiaoqi Zhao, Youwei Pang, Lihe Zhang, Huchuan Lu, and Lei Zhang.
\newblock Suppress and balance: A simple gated network for salient object detection.
\newblock In \emph{ECCV}, 2020{\natexlab{b}}.

\bibitem[Zhou et~al.(2023{\natexlab{a}})Zhou, Li, Loy, and Dai]{zhou2023edgesam}
Chong Zhou, Xiangtai Li, Chen~Change Loy, and Bo Dai.
\newblock Edgesam: Prompt-in-the-loop distillation for on-device deployment of sam.
\newblock \emph{arXiv preprint arXiv:2312.06660}, 2023{\natexlab{a}}.

\bibitem[Zhou et~al.(2022{\natexlab{a}})Zhou, Feng, Gu, Cheng, Lu, Shi, and Ma]{zhou2022uncertainty}
Qianyu Zhou, Zhengyang Feng, Qiqi Gu, Guangliang Cheng, Xuequan Lu, Jianping Shi, and Lizhuang Ma.
\newblock Uncertainty-aware consistency regularization for cross-domain semantic segmentation.
\newblock \emph{CVIU}, 2022{\natexlab{a}}.

\bibitem[Zhou et~al.(2022{\natexlab{b}})Zhou, Feng, Gu, Pang, Cheng, Lu, Shi, and Ma]{zhou2022context}
Qianyu Zhou, Zhengyang Feng, Qiqi Gu, Jiangmiao Pang, Guangliang Cheng, Xuequan Lu, Jianping Shi, and Lizhuang Ma.
\newblock Context-aware mixup for domain adaptive semantic segmentation.
\newblock \emph{TCSVT}, 2022{\natexlab{b}}.

\bibitem[Zhou et~al.(2022{\natexlab{c}})Zhou, Zhang, Yao, Yi, Ding, and Ma]{zhou2022adaptive}
Qianyu Zhou, Ke-Yue Zhang, Taiping Yao, Ran Yi, Shouhong Ding, and Lizhuang Ma.
\newblock Adaptive mixture of experts learning for generalizable face anti-spoofing.
\newblock In \emph{ACM MM}, 2022{\natexlab{c}}.

\bibitem[Zhou et~al.(2022{\natexlab{d}})Zhou, Zhang, Yao, Yi, Sheng, Ding, and Ma]{zhou2022generative}
Qianyu Zhou, Ke-Yue Zhang, Taiping Yao, Ran Yi, Kekai Sheng, Shouhong Ding, and Lizhuang Ma.
\newblock Generative domain adaptation for face anti-spoofing.
\newblock In \emph{ECCV}, 2022{\natexlab{d}}.

\bibitem[Zhou et~al.(2022{\natexlab{e}})Zhou, Zhuang, Yi, Lu, and Ma]{zhou2022domain}
Qianyu Zhou, Chuyun Zhuang, Ran Yi, Xuequan Lu, and Lizhuang Ma.
\newblock Domain adaptive semantic segmentation via regional contrastive consistency regularization.
\newblock In \emph{ICME}, 2022{\natexlab{e}}.

\bibitem[Zhou et~al.(2023{\natexlab{b}})Zhou, Gu, Pang, Lu, and Ma]{zhou2023self}
Qianyu Zhou, Qiqi Gu, Jiangmiao Pang, Xuequan Lu, and Lizhuang Ma.
\newblock Self-adversarial disentangling for specific domain adaptation.
\newblock \emph{TPAMI}, 2023{\natexlab{b}}.

\bibitem[Zhou et~al.(2023{\natexlab{c}})Zhou, Li, He, Yang, Cheng, Tong, Ma, and Tao]{zhou2022transvod}
Qianyu Zhou, Xiangtai Li, Lu He, Yibo Yang, Guangliang Cheng, Yunhai Tong, Lizhuang Ma, and Dacheng Tao.
\newblock Transvod: end-to-end video object detection with spatial-temporal transformers.
\newblock \emph{TPAMI}, 2023{\natexlab{c}}.

\bibitem[Zhou et~al.(2023{\natexlab{d}})Zhou, Zhang, Yao, Lu, Yi, Ding, and Ma]{zhou2023instance}
Qianyu Zhou, Ke-Yue Zhang, Taiping Yao, Xuequan Lu, Ran Yi, Shouhong Ding, and Lizhuang Ma.
\newblock Instance-aware domain generalization for face anti-spoofing.
\newblock In \emph{CVPR}, 2023{\natexlab{d}}.

\bibitem[Zhou et~al.(2024)Zhou, Zhang, Yao, Lu, Ding, and Ma]{zhou2024est}
Qianyu Zhou, Ke-Yue Zhang, Taiping Yao, Xuequan Lu, Shouhong Ding, and Lizhuang Ma.
\newblock Test-time domain generalization for face anti-spoofing.
\newblock In \emph{CVPR}, 2024.

\bibitem[Zhu et~al.(2020)Zhu, Su, Lu, Li, Wang, and Dai]{zhu2020deformable}
Xizhou Zhu, Weijie Su, Lewei Lu, Bin Li, Xiaogang Wang, and Jifeng Dai.
\newblock Deformable detr: Deformable transformers for end-to-end object detection.
\newblock In \emph{ICLR}, 2020.

\bibitem[Zhuge et~al.(2022)Zhuge, Fan, Liu, Zhang, Xu, and Shao]{zhuge2022salient}
Mingchen Zhuge, Deng-Ping Fan, Nian Liu, Dingwen Zhang, Dong Xu, and Ling Shao.
\newblock Salient object detection via integrity learning.
\newblock \emph{TPAMI}, 2022.

\bibitem[Zhuge et~al.(2023)Zhuge, Fan, Liu, Zhang, Xu, and Shao]{9789317}
Mingchen Zhuge, Deng-Ping Fan, Nian Liu, Dingwen Zhang, Dong Xu, and Ling Shao.
\newblock Salient object detection via integrity learning.
\newblock \emph{TPAMI}, 2023.

\end{thebibliography}
}


\end{document}